\documentclass{article}

\usepackage{arxiv}

\usepackage[utf8]{inputenc} 
\usepackage[T1]{fontenc}    
\usepackage{hyperref}       
\usepackage{url}            
\usepackage{booktabs}       
\usepackage{amsfonts}       
\usepackage{nicefrac}       
\usepackage{microtype}      
\usepackage{lipsum}

\usepackage{pgfplots}
\usepackage{pgfplotstable}
\usepackage{adjustbox}
\usepackage{enumitem}
\usepackage{tikz}
\usepackage{subfigure}
\usepackage{pgfplots}
\usepackage{boldline}
\usetikzlibrary{calc}

\usepgfplotslibrary{dateplot}
\pgfplotsset{compat=1.12}
\usetikzlibrary{fillbetween}

\usepackage{graphicx,xcolor}
\usepackage[linesnumbered,ruled,vlined]{algorithm2e}

\usepackage{float}
\restylefloat{table}

\SetCommentSty{mycommfont}

\newcommand{\useupquotes}{%
  \begingroup\lccode`\~=`\'\lowercase{\endgroup\let~}\algoupquote
  \begingroup\lccode`\~=`\"\lowercase{\endgroup\let~}\algoupquotes
  \catcode`\'=\active\catcode`\"=\active
}
\SetKwInput{KwInput}{Input}                
\SetKwInput{KwOutput}{Output}              
\SetKw{Continue}{continue}
\SetKw{True}{true}
\SetKw{False}{false}
\SetKw{Null}{null}
\SetKw{Andd}{and}
\SetKw{Return}{return}
\SetKw{Throw}{throw}
\SetKwProg{Init}{init}{:}{}
\SetKwProg{try}{try}{:}{}
\SetKwProg{catch}{catch}{:}{end}

\title{Ethanos: Lightweight Bootstrapping for Ethereum}

\author{
  Jae-Yun Kim\\
  Seoul National University\\
  \texttt{jaeykim@altair.snu.ac.kr} \\
  \And
  Jun-Mo Lee\\
  Seoul National University\\
  \texttt{jmlee@altair.snu.ac.kr} \\
  \And
  Yeon-Jae Koo\\
  Seoul National University\\
  \texttt{yjkoo@altair.snu.ac.kr} \\
  \And
  Sang-Hyeon Park\\
  Seoul National University\\
  \texttt{lukepark@altair.snu.ac.kr} \\
  \And
  Soo-Mook Moon\\
  Seoul National University\\
  \texttt{smoon@snu.ac.kr} \\
}

\begin{document}
\maketitle

\begin{abstract}
As ethereum blockchain has become popular, the number of users and transactions has skyrocketed, causing an explosive increase of its data size. As a result, ordinary clients using  PCs or smartphones cannot easily bootstrap as a {\em full node}, but rely on other full nodes such as the miners to run or verify transactions. This may affect the security of ethereum, so light bootstrapping techniques such as {\em fast sync} has been proposed to download only parts of full data, yet the space overhead is still too high. One of the biggest space overhead that cannot easily be reduced is caused by saving the state of all accounts in the block's {\em state trie}. Fortunately, we found that more than 90\% of accounts are inactive and old transactions are hard to be manipulated. Based on these observations, this paper propose a novel optimization technique called \textit{ethanos} that can reduce bootstrapping cost by {\em sweeping} inactive accounts periodically and by not downloading old transactions. If an inactive account becomes active, ethanos restore its state by running a restoration transaction. Also, ethanos gives incentives for \textit{archive nodes} to maintain the old transactions for possible re-verification. We implemented ethanos by instrumenting the \textit{go-ethereum (geth)} client and evaluated with the real 113 million transactions from 14 million accounts between $7M^{th}$ and $8M^{th}$ blocks in ethereum. Our experimental result shows that ethanos can reduce the size of the account state by half, which, if combined with removing old transactions, may reduce the storage size for bootstrapping to around 1GB. This would be reasonable enough for ordinary clients to bootstrap on their personal devices.
\end{abstract}


\section{Introduction} \label{sec:introduction}
Since the development of Bitcoin~\cite{nakamoto2008bitcoin}, ethereum has become one of the most popular blockchains because it introduced \textit{smart contract}, a tiny program with its own state and member functions to perform contract-related work. Because ethereum guarantees the validity of the smart contract state and its transitions, people began to issue their own digital assets called \textit{tokens} using smart contracts to raise funds, starting \textit{initial coin offering (ICO)} boom that makes ethereum even more popular. Unfortunately, the number of accounts and transactions soared explosively, making the ethereum data more than 3.0TB as of Nov 2019~\cite{etherscan_archivesync}. As a result, ordinary people using personal devices with limited storage cannot easily bootstrap as a \textit{full node}, a node that can fully verify  transactions and blocks, because they should first download the whole transactions and replay the transactions from the genesis block to verify the current block state while saving all intermediate block states. If they cannot fully validate transactions and blocks for themselves, they have to rely on other full nodes such as the miners or the central service providers to send and validate their transactions. However, relying on other nodes would compromise the security of a blockchain, because participants can be deceived to be a victim of financial fraud or their transaction could be censored by full nodes. What makes the matter worse, a small number of full nodes and miners are easier to manipulate the network, affecting the own spirit of the blockchain.

To solve the problem, we need to reduce bootstrapping cost to allow anyone to freely participate in the network. One primary bottleneck of bootstrapping is the \textit{state trie} that stores all histories of the account states by using a data structure named \textit{modified 
Merkle Patricia Trie}~\cite{fredkin1960trie,leis2013adaptive}. Actually, ethereum has tried to reduce the bootstrapping cost by providing a synchronization mode named \textit{fast sync} that downloads a snapshot of the \textit{pivot block}, current block - 64, and replays only the transactions from the pivot block to the current block to reproduce the current block state. This can obviate saving the state tries for those blocks older than the pivot block. Fast sync significantly reduces the storage size to 200GB as of Nov 2019~\cite{etherscan_fastsync}, which is only 6.7\% of the whole ethereum data, yet still too heavy to run on the personal devices. Moreover, the size of fast sync is continuously increasing because of the growing number of transactions and accounts. Even if we could reduce the download data further somehow (e.g., do not download old transactions made earlier than the pivot block since we can prove the past state of an account by requesting old transactions on demand from other full nodes \cite{personal-correspondence-with-Vitalik}), the state trie must hold all accounts for security reasons. For example, every account which has ever involved in any transaction must be stored in the current state trie to prevent \textit{replay attack}, an attack to re-transmit an authorized transaction multiple times to duplicate its execution; \emph{nonce} in an account that counts the transactions sent from the account can prevent such an attack.

\begin{figure}[t]
    \centering
    \includegraphics[width=0.5\linewidth]{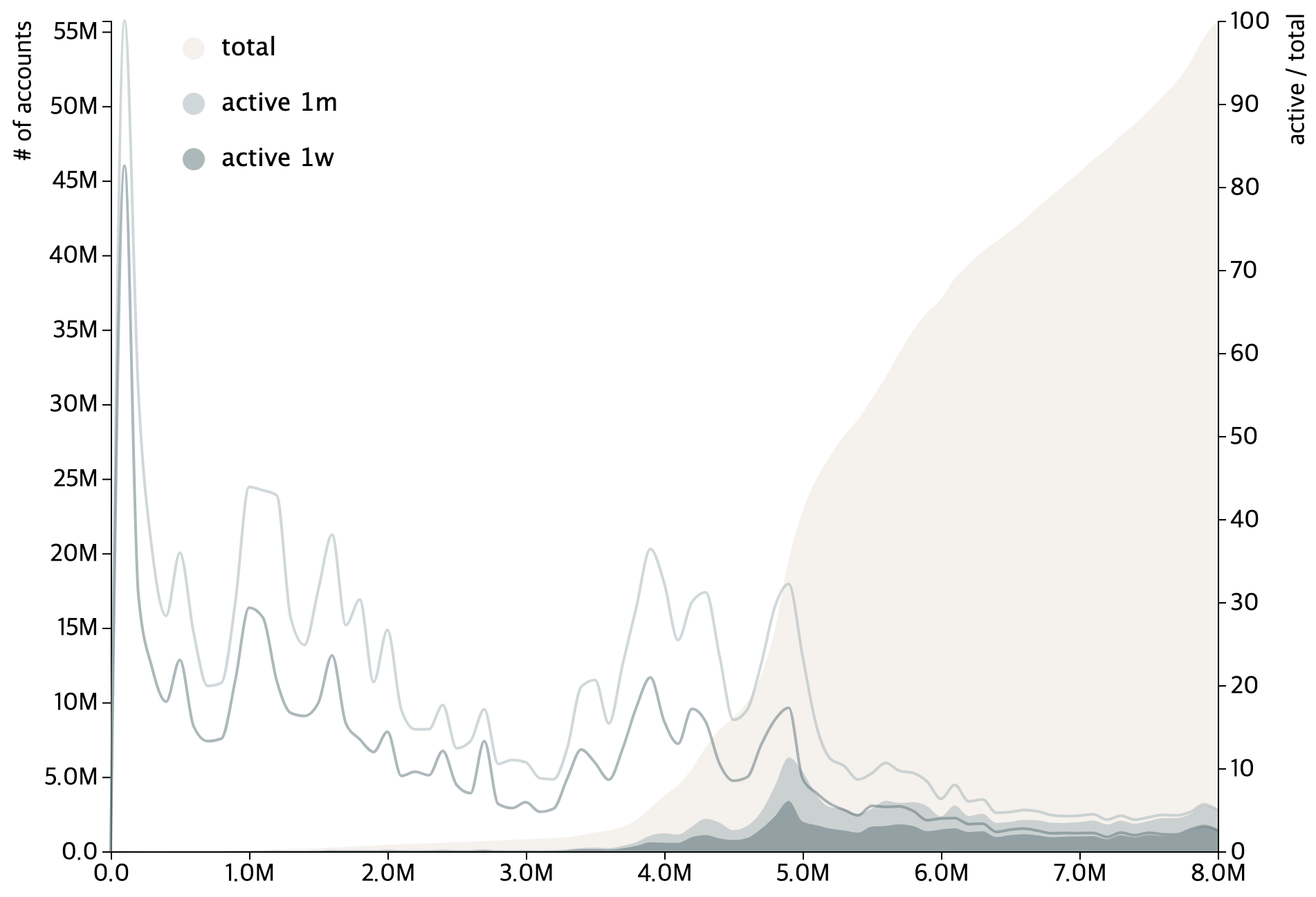}
    \caption{The ratio of the active accounts in ethereum}
    \label{fig:active_account_ratio}
\end{figure}

Above discussion indicates that the state trie of accounts is a bottleneck in bootstrapping cost reduction. Fortunately, we found that more then 95\% of the ethereum accounts are now inactive for a week or for a month from the current block and the ratio is increasing, when we trace the number of active accounts from the genesis block to the block number 8$M$, as depicted in Figure~\ref{fig:active_account_ratio}. This means that it is highly inefficient to hold all accounts in the state trie. To exploit this observation, we proposes a new optimization technique named \textit{ethanos} that sweeps inactive accounts periodically to reduce the size of the current state trie. Since it is not easy to separate and remove inactive accounts from the full state trie, ethanos simply build an empty trie at the start of each period and adds active accounts who made transactions during that period using the cached trie of the previous period. When an inactive account wants to invoke a transaction, it first invokes a restore transaction to reactivate itself, using Merkle proofs, void proofs, and bloom filters of previous periods. When we experimented with full sync for blocks between 7$M$ and 8$M$, ethanos could reduce the state trie size of ethereum by half. Actually, if we were not downloading the old transactions before the pivot block as discussed above, this result means that the bootstrapping size could be reduced to around 1GB. Although the download data would increase if we experiment with the full ethereum from the genesis block, we expect that the bootstrapping size would be limited to a few GBs, which seems to be highly promising for reducing the bootstrapping cost. In fact, our result also shows that the bootstrapping time is also reduced significantly. This paper made the following contributions:

\begin{itemize}
\item {We thoroughly analyzed the historical data of ethereum to produce statistics of the account states and analyze the effect of sweeping inactive accounts.}

\item {We designed the mechanism to sweep inactive accounts and to restore them with \textit{restore transaction}.}

\item {We evaluated our work with real ethereum transactions recorded from $7M+1$ to $8M$ blocks to show that we can reduce the bootstrapping cost.}
\end{itemize}

The rest of this paper are as follows. Section~\ref{sec:ethereum} reviews some background concepts of ethereum to make the paper self-contained. Section~\ref{sec:motivation} discusses minimum essential data required for bootstrapping as a full node. Section~\ref{sec:ethanos} describes the overall design of ethanos and its detailed mechanism of sweeping and restoration. The evaluation results are in Section~\ref{sec:evaluation} and related work is in Section~\ref{sec:related_work}. Summary and future work is discussed in Section~\ref{sec:summary}.

\begin{figure*}[t]
  \includegraphics[width=\linewidth]{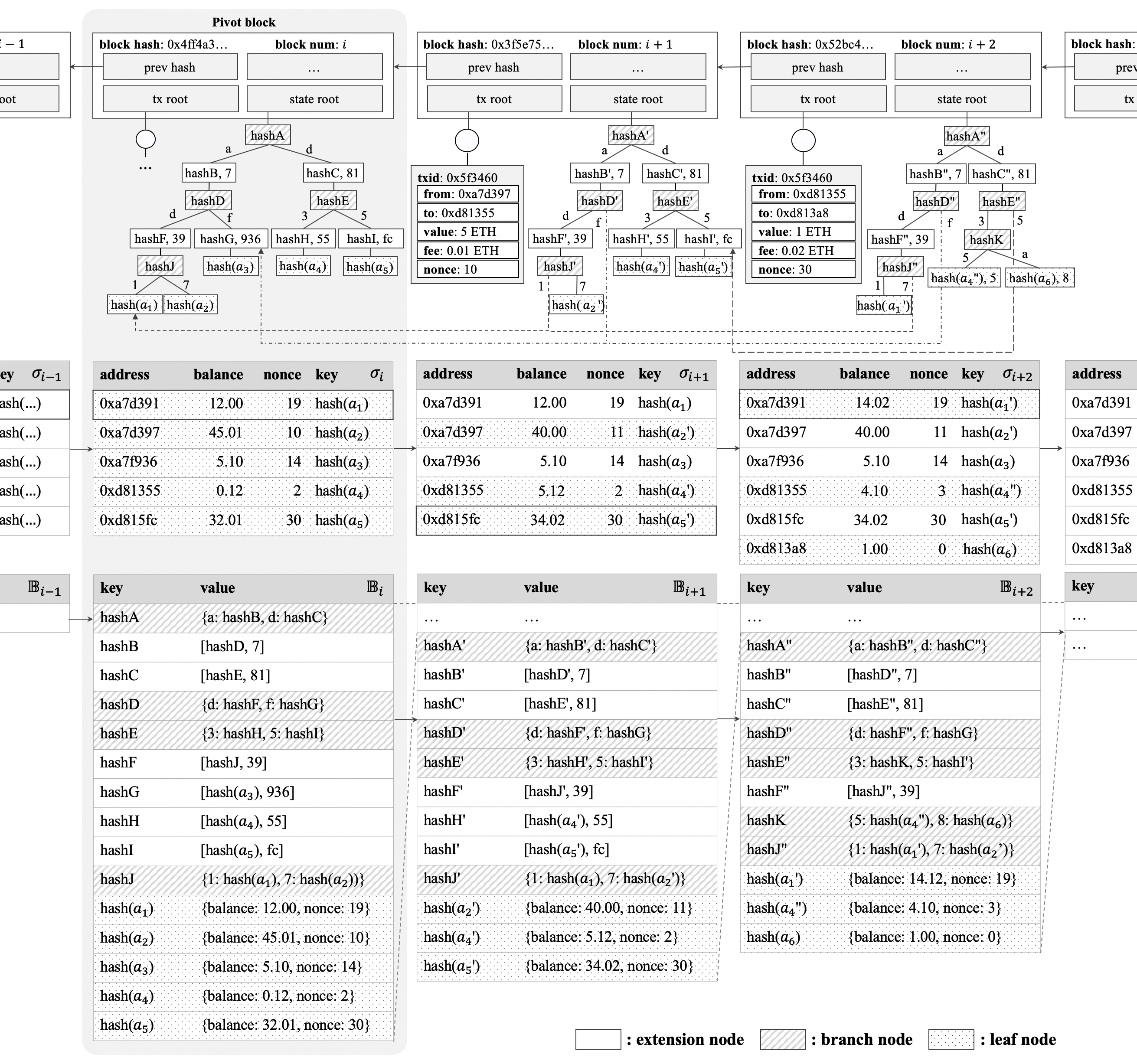}
  \caption{Simplified state and database structure of the ethereum.}
  \label{fig:ethereum}
\end{figure*}

\section{Ethereum} \label{sec:ethereum}
Ethereum is a \textit{transaction-based state machine} that maintains a single state among the nodes in the p2p network. A state~($\sigma$) is a set of address-account pairs, and they are updated by transactions~\cite{wood2014ethereum}. The updated state and the transactions are recorded into a block and propagated across the entire ethereum network by \textit{gossip protocol}~\cite{demers1988epidemic}. Block is a unit of state synchronization created every round, and a node who make a block is called \textit{miner}. Others receive the block and replay the transactions to reproduce and verify the current state. When most participating nodes synchronize the block with the current state $\sigma_i$, a new round is started to generate the next state $\sigma_{i+1}$. A miner is determined by sortition using \textit{Ethash}~\cite{ethash} similar to \textit{Proof-of-Work} in Bitcoin~\cite{nakamoto2008bitcoin} to implement pseudo-random, and uses \textit{GHOST protocol}~\cite{sompolinsky2015secure} similar to \textit{longest chain rule} in Bitcoin to solve the race condition when more than two blocks are generated in the same round. These algorithms are called \textit{consensus algorithm} that ensures all participants synchronize to the identical current block.

\subsection{Account}
Account in ethereum is similar to bank account, accessed by its {\em address} composed of 40 hexadecimal characters. An account includes 4 types of data: \emph{nonce}, \emph{balance}, \emph{storageRoot} and \emph{codeHash}. \emph{Nonce} is for keeping the transaction orders, and the nonce of an account in the current state is the number of all send transactions that the account has ever made. \emph{Balance} shows the amount of Ether (ETH), the asset of ethereum, that the account has. If an account is \textit{externally owned account (EOA)}, \emph{storageRoot} and \emph{codeHash} are \textit{null}. However, if an account is \textit{contract account (CA)}, \emph{storageRoot} shows the key of the storage where the value is the state of the smart contract, and \emph{codeHash} represents the hash value of the code used to retrieve and validate the code from the database.

\subsection{Transaction}
Transaction transfers a value from a sender account (\emph{from}) to a recipient account (\emph{to}), changing the state of both accounts. Transaction not only updates the balances, but also increments the nonce of the sender account. To prove validity of a transaction, we need to check if the transaction has the same nonce as the sender account's. A transaction includes a transaction fee to be taken by the miner. Detailed transaction structures in ethereum is more complicated, so we skip them for simplicity.

\subsection{State Transition}
Let $Tx_i$ be a transaction list of the $i^{th}$ block, and $Tx_i^n$ be the $n+1^{th}$ transaction of the $Tx_i$ list~\cite{wood2014ethereum, al2018fraud}. 
Formally:
\begin{equation} 
    Tx_i = [Tx_i^0, Tx_i^1, \cdots, Tx_i^{j-1}]
    \label{equation:tx_list}
\end{equation}
Then, state transition is represented as:
\begin{equation} 
    \sigma_i = \gamma(\sigma_{i-1}, Tx_i)
    \label{equation:state_transition}
\end{equation}
where $\sigma_i$ is $i^{th}$ state, and state transition function $\gamma$ applies each transaction in $Tx_i$ to the previous state $\sigma_{i-1}$. In the same manner, we can represent $i^{th}$ state $\sigma_i$ by sequentially applying transaction lists to the genesis state $\sigma_0$:
\begin{equation}
    \sigma_i = \gamma(\cdots\gamma(\gamma(\sigma_0, Tx_1),Tx_2),\cdots,Tx_i)
    \label{equation:current_state}
\end{equation}
For example, in Figure~\ref{fig:ethereum}, $i^{th}$ block state $\sigma_i$ consists of 5 accounts. We represented addresses as 6 characters and assumed that all accounts are EOA for the simplicity. If the transaction list $Tx_{i+1}$ contains one transaction $Tx_{i+1}^0$ that transfers 5 ETH with 0.01 ETH transaction fee from account 0xa7d397($a_2$) to 0xd81355($a_4$), the balance of 0xa7d397($a_2$) decreases from 45.01 to 40.00 ETH and the nonce is increased by one. On the other hand, the balance of 0xd81355($a_4$) increases from 0.12 to 5.12 ETH, but nonce does not change, because nonce counts only the sender's transactions. Meanwhile, 0xd815fc($a_5$) is the miner of the ${i+1}^{th}$ round, so that the balance of 0xd815fc($a_5$) is incremented by 2.01 ETH where 0.01 ETH is the transaction fee and 2 ETH is the block reward.

\subsection{State Trie}
Ethereum stores histories of accounts for each block by creating state trie which uses a data structure named \textit{modified merkle patricia trie (MPT)}~\cite{wood2014ethereum,merkle-patricia-tree} that combines patricia trie with merkle tree. Basically, patricia trie (or radix trie)~\cite{fredkin1960trie,leis2013adaptive} arranges (\emph{key}, \emph{value}) bindings with the character sequence of the key to make it a deterministic path of the value. Merkle tree (or hash tree)~\cite{merkle1987digital} labels every leaf node with the hash of the values, and labels every intermediate node with the hash of the labels of its child nodes, which provides efficient and secure verification of the contents of large data structure~\cite{merkletree}. MPT places accounts as leaf nodes and generates parent nodes with the common prefixes to make each address a path for the account. Intermediate nodes are composed of \textit{extension node} and \textit{branch node}, which are labeled by hash of the node values, and stored to the key-value database like levelDB~\cite{leveldb}. Extension node is a 2-item node of the form [\emph{path}, \emph{key}] where \emph{path} is the common prefixes of the child nodes, and the \emph{key} is for the next db lookup~\cite{merkle-patricia-tree}. Branch node is a 17-item node of the form [\emph{v0}, \emph{v1}, $\cdots$, \emph{v15}, value] where index 0 to 15 represents the 1 character common prefix of the child nodes, and elements at each index are the keys for the next db lookup. Figure~\ref{fig:ethereum} shows a precise implementation of the MPT, which is composed of 5 accounts($a_{1\sim5}$) in the $i^{th}$ block of which states are combined to make $\sigma_i$. We represent hash values of each record in db "hash$\ast$", and branch nodes only with indexes with values for simplicity. Since the keys are cryptographic hash of the node values, each time a leaf node is updated, all nodes in that path are created and inserted into the database, which causes an explosion in state trie size.

\subsection{Proof of Membership and Non-Membership}
Block header stores the summary of the block data including \emph{prev hash}, \emph{tx root} and \emph{state root}. \emph{tx root} is the root hash value of the transaction trie and \emph{state root} is the root hash value of the state trie. The membership of a transaction or an account can be proved with block header and data named \textit{merkle proof} composed of partial nodes of the trie. For example, at the $i^{th}$ block in Figure~\ref{fig:ethereum}, prover can prove the membership of account 0xd81355($a_5$) with merkle proof:
\begin{equation} 
P_M(a_5) = \mathbb{B}([hashA, hashC, hashE, hashI, hash5])
\end{equation}
A verifier can reproduce hash keys by applying hash function to the values in database $\mathbb{B}$ like
\begin{equation} 
hashA = hash(\{a: hashB, d: hashC\})
\end{equation}
from the merkle proof and finally compares $hash(hashA)$ with \emph{state root} to determine the membership of the account. In the same way, we can prove non-membership of an account by generating void proof. For example, a prover can prove non-membership of 0xa7ec6b($a_n$) in $i^{th}$ state $\sigma_i$ by generating void proof:
\begin{equation} 
P_V(a_n) = \mathbb{B}([hashA, hashB', hashD'])
\end{equation}
A verifier can reproduce $hashA$ to verify the state root, and determine that
$\mathbb{B}(hashD)=\{d: hashF, f: hashG\}$ do not contain the key $e$, which implies the non-membership.

\subsection{Bootstrapping}
When a new node wants to join the ethereum network as a full node, it must synchronize to a valid current state using the \textit{block headers} and the \textit{transactions in each block}. So, the \textit{minimum essential data (MED)} are as follows ~\cite{qian2018improved}:
\begin{itemize}
    \item Block headers
    \item Transactions in each block
    \item Current state
\end{itemize}
A node reproduces the current state by replaying the transactions from the genesis block to the current block. If a node eventually catches up with the current state, bootstrapping process is finished and the node becomes a full node that can validate a new transaction. This is called \textit{full sync}. However, due to the enormous data of the blockchain, full sync requires huge storage and long running time to complete the bootstrapping process. Therefore, geth client provides another bootstrapping mode named \textit{fast sync}, which downloads the state trie of a \textit{pivot block} that is 64 blocks ahead of the current block, and replays the transactions between the two blocks to reproduce the current state. In this way, fast sync can significantly reduce the storage size and the bootstrapping time. Although there are much fewer transactions to replay, they cannot be replayed until the state trie of the pivot block is completely downloaded. We found that the downloading time is huge due to the frequent disk I/O in the key-value storage, comprising a bottleneck of bootstrapping. Fast sync still requires a large storage size about 200GB as of Nov 2019, which is continuously growing.

\section{Observation and Insight for Ethanos} \label{sec:motivation}
The goal of ethanos is to drastically reduce the bootstrapping cost by making fast sync more efficient, so as to allow the ordinary clients to fully verify and propagate transactions on their own. To achieve this goal, we examined 
the MED of ethereum, based on the two aspects as was done for Bitcoin \cite{leung2019vault}: \textit{width}, which is the current state, and \textit{length}, which is the chain of block headers and the transactions
for each block.

Width is essential data to verify a new transaction, which is the current state $\sigma_c$ in ethereum. When one makes a transaction to transfer some value to someone else in the current state $\sigma_c$, the miner verifies it with $\sigma_c$ to check if the balance of the sender is greater then the transferring value: $\sigma_c[sender]_{balance} >= value$.
Miner also verifies \textit{nonce}, a transaction count of the sender, is same as that of the transaction: $\sigma_c[sender]_{nonce} == Tx_{nonce}$.

\begin{figure}[t]
\centering
\def\angle{90}
\def\radius{3}
\def\cyclelist{{"orange","red","green","blue","yellow","gray","violet"}}
\newcount\cyclecount \cyclecount=-1
\newcount\ind \ind=-1\
\begin{adjustbox}{width=.4\textwidth}
\begin{tikzpicture}[nodes = {font=\sffamily}]
  \foreach \percent/\name in {
      62.34/1 day,
      13.66/1 week,
      6.38/2 weeks,
      6.57/1 month,
      9.45/6 months,
      1.60/others
    } {
      \ifx\percent\empty\else               
        \global\advance\cyclecount by 1     
        \global\advance\ind by 1            
        \ifnum6<\cyclecount                 
          \global\cyclecount=0              
          \global\ind=0                     
        \fi
        \pgfmathparse{\cyclelist[\the\ind]} 
        \edef\color{\pgfmathresult}         
        \draw[fill={\color!50},draw=white] (0,0) -- (\angle:\radius)
          arc (\angle:\angle+\percent*3.6:\radius) -- cycle;
        \node at (\angle+0.5*\percent*3.6:0.7*\radius) {\percent\,\%};
        \node[pin=\angle+0.5*\percent*3.6:\name]
          at (\angle+0.5*\percent*3.6:\radius) {};
        \pgfmathparse{\angle+\percent*3.6}  
        \xdef\angle{\pgfmathresult}         
      \fi
    };
\end{tikzpicture}
\end{adjustbox}
\caption{Average transaction distances of the accounts}
\label{fig:average_account_frequency}
\end{figure}
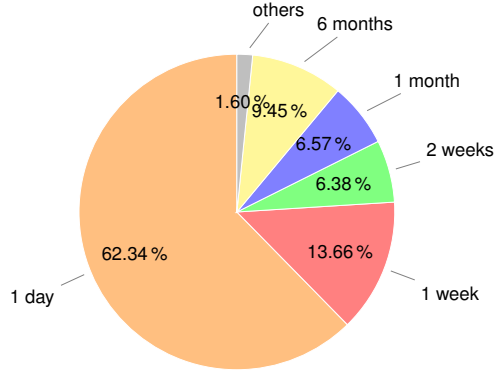

The most dominant part of $\sigma_c$ is the state trie. Since the number of accounts in ethereum is about 79 million as of Nov 2019 and is continuously growing about 50K-150K every day~\cite{etherscan_address}, they would make a seriously wide state trie, thus taking a long time to download. Moreover, such a state trie would have a long height, which makes the replay of transactions take time since the change of accounts need to update the hash values of all nodes in the MPT paths.

Fortunately, we found that most accounts are inactive after short periods of trading by inspecting the transactions in ethereum from 1 to $8M$ block as previously depicted in Figure~\ref{fig:active_account_ratio}. 
By inspecting the average distance between the blocks when an account is updated (when it is a sender, a recipient of a transaction, or a miner of a block), we classified the update period as shown in Figure~\ref{fig:average_account_frequency}. We found that 76\% of the account states were updated in a week ($\approx$40,320 blocks) and about 90\% of the account states were updated in a month ($\approx$172,800 blocks). Considering that the active account ratios during a week and a month are under 5\% in Figure~\ref{fig:active_account_ratio}, we can expect that eliminating inactive accounts once a week or once a month would significantly reduce the width, without compromising user experience that user have to send a restore transaction before trading.

Length is essential data to verify the current state, which are the block headers and the transactions in each block. The current state is the execution result of all transaction history as depicted at equation~\ref{equation:current_state}. However, given the nature of blockchain that older transactions are more hard to fabricate, we can eliminate old transactions by incentivizing archive nodes to maintain the entire transactions and response to transaction requests. We can simply implement a bootstrapping mode that does not downloads the transactions before the pivot block(current block - 64) to confine the length, and we named it \textit{compact sync}. However, compact sync has to downloaded whole block headers to validate old transactions when needed, which is very small that it is negligible.

\section{Design and Implementation of Ethanos} \label{sec:ethanos}
Previous section indicated that there are many inactive accounts in ethereum, which would better be swept to reduce the size of the state trie. This section describes the design and implementation of ethanos, a new ethereum-based blockchain for sweeping and restoration of accounts.

\subsection{Overall Design}
Ethanos is an optimization technique for ethereum, yet sharply reducing bootstrapping cost by reducing the current state and transactions to download. The current state is reduced by composing the current state only with active accounts by sweeping dormant accounts periodically, which is backed up by restoration mechanism to guarantee secure reactivation of dormant accounts on demand. Transactions are eliminated when bootstrapping by not downloading transactions in old blocks, instead, incentivize a certain amount nodes to maintain the whole transaction histories in exchange for fee of restore transactions that only archive nodes can make. We call a node maintaining the whole blockchain \textit{archive node} and a node lightly bootstraps from the archive node \textit{compact node}.

\begin{figure*}[t]
    \centering
    \includegraphics[width=\linewidth]{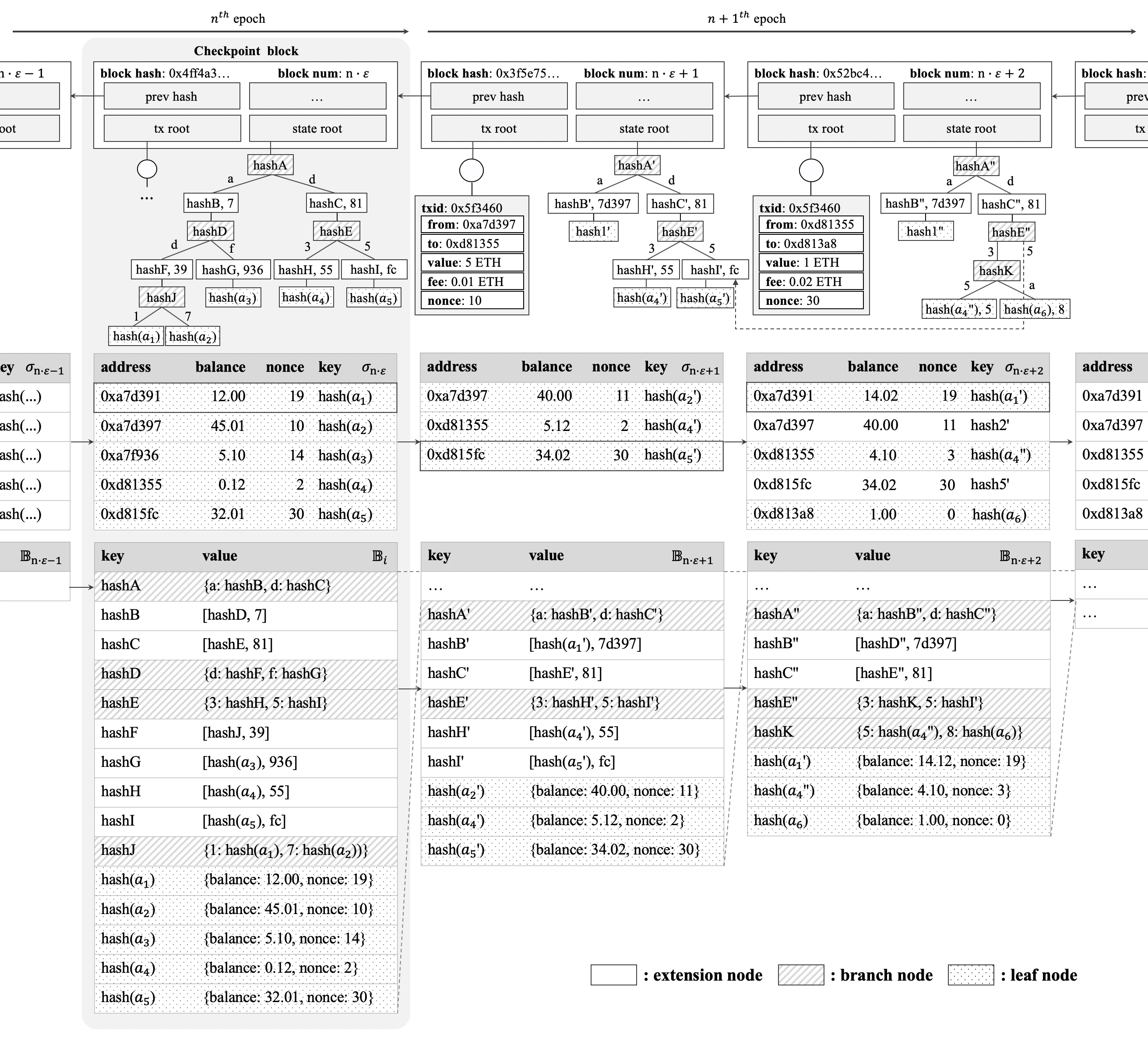}
    \caption{Simplified state and database structure of the ethanos.}
    \label{fig:ethanos}
\end{figure*}

\subsection{Sweep} \label{subsec:sweep}
Sweep process periodically removes dormant accounts from the latest state trie by a certain epoch $\epsilon$, and the last block of each epoch is a \textit{checkpoint}. However, it is very expensive to traverse a state trie to find and remove dormant accounts, so we designed \textit{sweep} mechanism that creates a new empty state trie every time a new epoch starts and maintains the accounts active during the new epoch. Meanwhile, the last checkpoint is cached to serve the latest state of an account in case it is not in the current state. For example in Figure~\ref{fig:ethanos}, $n^{th}$ epoch ends at the $n\cdot\epsilon$ block, and $n+1^{th}$ epoch starts with an empty state $\sigma_\phi$. A miner 0xd815fc($a_5$) wants to execute a transaction transferring value from 0xa7d397($a_2$) to 0xd81355($a_4$) to create $n\cdot\epsilon+1$ block, but the current state $\sigma_{n\cdot\epsilon+1}$ does not have the any accounts because it is newly created as an empty state $\sigma_\phi$. Therefore, the miner searches the last checkpoint state $\sigma_{n\cdot\epsilon}$ to validate the transaction and update the current state $\sigma_{n\cdot\epsilon+1}$ by adding mining reward (2 ETH) and transaction fee (0.01 ETH) to itself, subtracting transferring value with fee from the \emph{from} address 0xa7d397($a_2$), and adding the transferring value to the \emph{to} address 0xd81355($a_4$). Now, the transaction transferring value from 0xd81355($a_4$) could be verified with the previous block state $\sigma_{i+1}$. However, the recipient 0xd813a8($a_6$) is not in both previous state $\sigma_{i+1}$ and the last checkpoint $\sigma_{n\cdot\epsilon}$. In this case, 0xd813a8($a_6$) is treated as a new account, which is the way ethereum creates a new account. In this manner, we can obtain a state trie consisting only of accounts active during the $n+1^{th}$ epoch at the next checkpoint $\sigma_{n\cdot(\epsilon+1)}$.

\subsection{Reactivation} \label{subsec:reactivation}
Ethanos guarantees secure reactivation of a dormant account by restoration process to bring back the latest state of the account. However, because of the way ethereum creates accounts~\ref{subsec:sweep}, a dormant account can be activated as a new account, which we call \emph{pawn account}. We will call the reactivation as a pawn account \textit{respawn} to distinguish from restoration.

\subsubsection{Restoration}
Restoration process is triggered by \textit{restore transaction} including essential data of which \emph{to} address is a specific address (e.g. 0x$0\cdots01234$). However, the restore transaction must be sent by another account, since the state of dormant account is not in the current state so that it cannot pay a transaction fee for restoration. The essential data restore transaction has to include is as follows:
\begin{itemize}
    \item  A merkle proof of the account state at the last active checkpoint
    \item  Void proofs of the account between the last active checkpoint and the latest checkpoint
\end{itemize}
Suppose a user wants to restore an account $a$ whose last active checkpoint is $k$, and the latest checkpoint number is $n$ where $k < n$. In this case, the state of the checkpoint $k$ contains the state of this account $\sigma_{k\cdot\epsilon}[a]$, while the states after the checkpoint $k$ do not have the account state $\sigma_{>k\cdot\epsilon}[a]=\phi$. So, the restore transaction includes the merkle proof $P_M[k\cdot\epsilon]_a$ to prove that the account state to restore is valid, and void proofs $M_V[>k\cdot\epsilon]_a$ to prove that the checkpoint $k$ is the last active checkpoint, Otherwise, a malicious user can recover an account state before the last active checkpoint $k$. However, as the $n - k$ increases, the size of the restore transaction also increases, which would cause a big burden. So we introduce a space-efficient probablistic data structure named \textit{bloom filter} to each block, which indicates a membership of an element~\cite{bloom1970space}. Bloom filter always answer positive for membership, but sometimes, also answers positive for the non-membership with a low probability (false positive). However, if a bloom filter answers negative, we can be sure for the non-membership. Therefore, when a bloom filter of a checkpoint $i$ answers negative for the address $a$ to restore, the restore transaction does not have to include void proof $P_V[i]_a$.

We add the 10MB bloom filter for each block to insert 

to each block header to validate  By introducing bloom filter, we can sharply reduce the number of void proofs and the size of restore transaction.

\subsubsection{Respawn}
A state of an inactive account could be reactivated by a normal transaction that transfers a value to that account. We named an account reactivated by a normal transaction \textit{pawn}, and the process \textit{respawn}. Respawn process is the same as the way a new account is created in ethereum, and the pawn can also issue a transaction if it is valid without restoration. So, there is no way to tell if the account is restored or respawned until seeing the checkpoint history.
However, if we treat a pawn account as a new account to set the \emph{nonce} as 0, an attacker can retrieve the balance by retransmitting a transaction already executed in the past (replay attack). Therefore, we set the initial nonce as the \textit{current block number} (k) times \textit{Maximum Transactions Per Account Per Block} that a nonce can hardly catch up with to eliminate the chance of replay attack~\cite{eip169}:
\begin{equation}
    \sigma_k[a]_{nonce} = k * C_{MAX\_TXS\_PER\_ACCT\_PER\_BLOCK}
    \label{equation:merge_nonce}
\end{equation}
Then, how can we restore an account with pawns?
First, nodes have to determine whether a reactivated state is a restored or respawned. we solve the problem by adding 1-bit restoration flag to the account state of which value is 1 when an account is restored, otherwise 0.
Second, merkle proofs of every checkpoint where a pawn account was active are required to determine the last state of the pawn account. If there are many pawn accounts in the block history from the last active checkpoint to the current block, all the merkle proofs are required to discover last state of each pawn account for valid restoration.
Third, last state of the original account and last states of each pawn accounts should be merged into a final state. We define the merge function $\Phi$ that merges balance and nonce of the states. For example, if we wants to merge account states of \emph{a} in block $i$ and $j$ when the current block number is $k$, $\Phi$ function is defined as:
\begin{equation} 
    \sigma_k[a] = \Phi(\sigma_i[a], \sigma_j[a])
    \label{equation:merge_function}
\end{equation}
The merged balance and nonce are addition of the two accounts.
\begin{equation} 
    \sigma_k[a]_{balance} = \sigma_i[a]_{balance} + \sigma_j[a]_{balance}
    \label{equation:merge_balance}
\end{equation}
\begin{equation} 
    \sigma_k[a]_{nonce} = \sigma_i[a]_{nonce} + \sigma_j[a]_{nonce}
    \label{equation:merge_balance}
\end{equation}

\subsection{Bootstrapping}
Ethanos has two bootstrapping modes, one is \textit{full archive sync} and the other is \textit{compact sync}. We call a node to bootstrap \textit{client}, and a node providing data to help bootstrapping of client \textit{host}. Full archive sync downloads all data from the host and validates and replays the transactions to reproduce the current state on the client, which is a replication of host. Compact sync is similar to \textit{fast sync} that downloads a state trie of the pivot block and replays the transactions after the pivot block to the current block. However, compact sync does not download the transactions before the pivot block unlike fast sync, which significantly reduces the bootstrapping cost. Also, the way to select a pivot block in ethanos is different from that of ethereum, because ethanos requires the state trie of the last checkpoint block to be a full node that is able to validate a new transaction. Therefore, it is advantageous to select the pivot block as the last checkpoint block. However, the overhead increases when the current block moves away from the last checkpoint block. So if the distance between the current block and the last checkpoint block, client can choose to synchronize from the last checkpoint or synchronize from the original pivot block, 64 block ahead of the current block, and downloads the last checkpoint state trie separately, which is 800MB on average.

\subsection{Incentives}
Ethanos reduces bootstrapping cost by periodically eliminating inactive accounts and old transactions. However, a certain number of archive nodes reserving the eliminated data are required to make restore transactions and provide old transactions for integrity and security of the blockchain. We can achieve the goal by rewarding them for creating a restore transaction, because old transactions are necessary to re-establish old states to create a valid restore transaction. And users are willing to pay fees to reinstate their accounts if the accounts are more valuable than the fee. The number of archive nodes and the fee will be determined by the principle of supply and demand. If the number of archive nodes decreases, the fee will be increased to attract more nodes to be archive nodes, and vice versa.

\begin{figure}[t]
    \centering
  \includegraphics[width=0.5\linewidth]{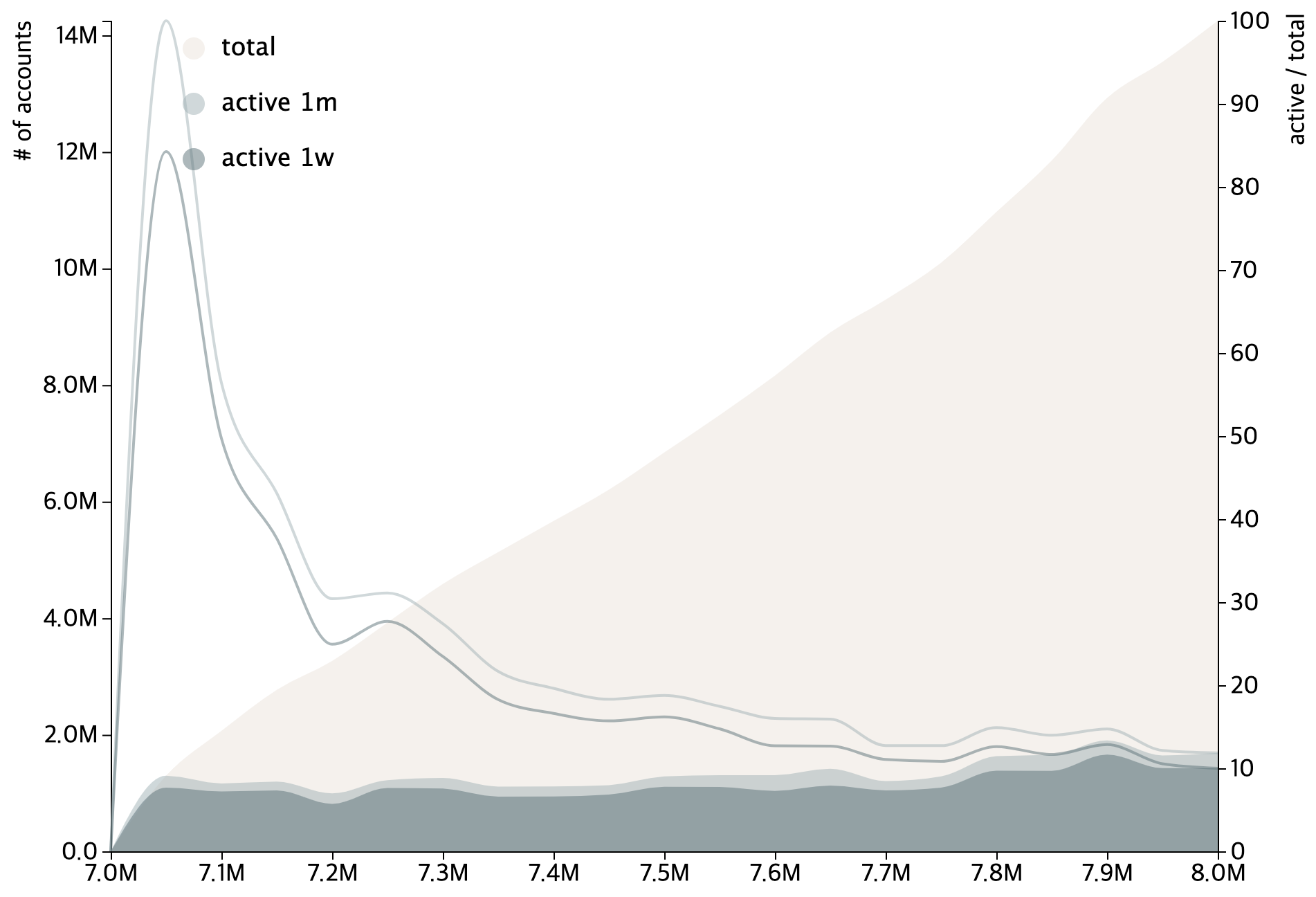}
  \caption{The ratio of the active accounts in ethereum for $7M+1$ to $8M$ blocks}
  \label{fig:active_account_ratio_7m}
\end{figure}

\section{Evaluation} \label{sec:evaluation}
The evaluation was performed on blocks and transactions from $7M+1$ to $8M$ blocks in ethereum with 114,973,077 transactions related with 14,264,037 accounts. As you can see in Figure~\ref{fig:active_account_ratio_7m}, active accounts are about 1 million and the ratio is reduced to 10\% at the $8M$ block. Therefore, we can expect that the bootstrapping cost of ethanos would be kept within a certain range, so that the ratio will be getting efficient over time. But we have to use 3 tricks to use the real transactions.
\begin{itemize}
    \item We made all transactions as \textit{delegated transactions} because we cannot make signatures of each transaction, which increases each transaction size about 20 Bytes.
    \item Because we set the state of the $7M+1$ block as an empty state, we made the value of each transaction as 0 to avoid exceptions incurred by insufficient balances.
    \item We treated \textit{contract accounts (CAs)} as \textit{externally owned accounts (EOAs)}~\cite{vogelsteller2014ethereum}, because we do not handle smart contract in this paper.
    \item \textit{Internal transactions} that a method of a smart contract transmits to another smart contract are not considered for the same reason.
    \item We relieved both size and gas limit of a transaction to insert merkle and void proofs to a restore transaction.
\end{itemize}
In addition, we assumed that every sweeped account sends restore transactions before transmitting transactions in order to make all transactions valid. We set the epoch as $17,230$ blocks (1 month), so the sweep processes are conducted 5 times during the experiment, and we compared our work with geth v1.9 client.  The experiments are performed on Ubuntu 18.04.3 LTS with AMD Ryzen Threadripper 1950X 16-Core @ 3.40GHz processor and 128GB memory with 1TB SSD storage.

\subsection{Storage size}
\subsubsection{Full Archive Sync}
Full archive sync is a sync mode that downloads and replays all the transactions from the genesis block to the current block reproducing the whole state histories of the blockchain. Table~\ref{table:sync_storage} gives full archive sync size of each data type for geth and ethanos at the $8M$ block. Ethanos reduces the total blockchain size about 16GB compared to geth. Especially, it reduces the size of \emph{Trie nodes} about 18GB, because ethanos only maintains the state trie only with active accounts during a month which are only 10\% of the total accounts. However, hash field of bloom filter in block header slightly increases \emph{Headers}, and restore transactions increment \emph{Bodies} and \emph{Receipts} about 1.6GB. As a result, ethanos reduced full archive node by about 8\% compared to ethereum.

\begin{table}[t]
\centering
\caption{Storage size(MB) comparison of \textit{full archive sync} between \emph{geth} and \emph{ethanos} from $7M+1$ to $8M$ block.}
\label{table:sync_storage}
\begin{adjustbox}{width=0.5\linewidth}
\footnotesize
\begin{tabular}{lrrr}
\hlineB{2}
Data type                           & geth        & ethanos       & Diff        \\ \hline
Headers                             & 316.68      & 348.16        & +31.48      \\
Bodies                              & 11,110.00   & 12,750.00     & +1,640.00   \\
Receipts                            & 3,578.97    & 3,605.66      & +26.69      \\
Difficulties                        & 15.57       & 15.33         & -0.24       \\
Block number -\textgreater{} hash   & 39.66       & 39.35         & -0.31       \\
Block hash -\textgreater{} number   & 39.10       & 39.10         & 0           \\
Transaction Index                   & 3,650.00    & 3,650.00      & 0           \\
Bloombit index                      & 20.49       & 20.49         & 0           \\
Trie nodes                          & 183,100.00  & 165,140.00    & -17,960.00  \\
Trie preimages                      & 857.00      & 888.11        & +31.11      \\ \hline
total                               & 202,727.47  & 186,496.20    & -16,231.27  \\
\hlineB{2}
\end{tabular}
\end{adjustbox}
\end{table}

\begin{table}[t]
\centering
\caption{Storage size(MB) comparison of \textit{fast sync} and \textit{compact sync} between \emph{geth} and \emph{ethanos} from $7M+1$ to $7M+864K$ ($5^{th}$ checkpoint) block.}
\label{table:sync_storage_fast_compact}
\begin{adjustbox}{width=0.5\linewidth}
\footnotesize
\begin{tabular}{lrrrr}
\hlineB{2}
                                    & \multicolumn{2}{c}{fast sync} & \multicolumn{2}{c}{compact sync} \\
                                    \cmidrule(lr){2-3}                          \cmidrule(lr){4-5}
Data type                           & geth        & ethanos     & geth          & ethanos\\ \hline
Headers                             & 276.66      & 303.83      & 276.66        & 303.83    \\
Bodies                              & 9,260.00    & 10,130.00   & 8.12          & 13.02     \\
Receipts                            & 2,973.66    & 2,979.81    & 10.43         & 10.44     \\
Difficulties                        & 11.49       & 11.48       & 11.49         & 11.48     \\
Block number -\textgreater{} hash   & 31.86       & 31.88       & 31.86         & 31.88     \\
Block hash -\textgreater{} number   & 33.79       & 33.79       & 33.79         & 33.79     \\
Transaction Index                   & 3,020.00    & 3,020.00    & 0.19          & 0.19      \\
Bloombit index                      & 0           & 0           & 0             & 0         \\
Trie nodes                          & 2,220.00    & 804.79      & 2,220.00      & 802.33    \\
Trie preimages                      & 0           & 0.28        & 0             & 0         \\ \hline
total                               & 17,827.46   & 17,315.86   & 2,592.54      & 1,206.97  \\
\hlineB{2}
\end{tabular}
\end{adjustbox}
\end{table}

\begin{figure}
\centering
\begin{tikzpicture}
\begin{axis}[
    axis lines=middle,
    ymin=0,
    ymax=23000,
    xmin=0.9,
    xlabel=\# of checkpoint,
    ylabel=size,
    x label style={at={(axis description cs:0.5,-0.05)},anchor=north},
    y label style={at={(axis description cs:-0.1,.5)},rotate=90,anchor=south},
    legend style={at={(0.03,0.7)},anchor=west},
    enlargelimits = false,
    xticklabels from table={storage.dat}{epoch},xtick=data]
\addplot[blue!50,thick,mark=square*] table [y=geth-fast,x=X]{storage.dat};
\addlegendentry{geth-fast}
\addplot[red!50,thick,mark=square*] table [y=ethanos-fast,x=X]{storage.dat};
\addlegendentry{ethanos-fast}
\addplot[green!50,thick,mark=square*] table [y=geth-compact,x=X]{storage.dat};
\addlegendentry{geth-compact}
\addplot[orange!50,thick,mark=square*] table [y=ethanos-compact,x=X]{storage.dat};
\addlegendentry{ethanos-compact}
\end{axis}
\end{tikzpicture}
\caption{Size of \textit{fast sync} and \textit{compact sync} between \emph{geth} and \emph{ethanos} at each checkpoint}
\label{fig:sync_size_of_each_checkpoint}
\end{figure}
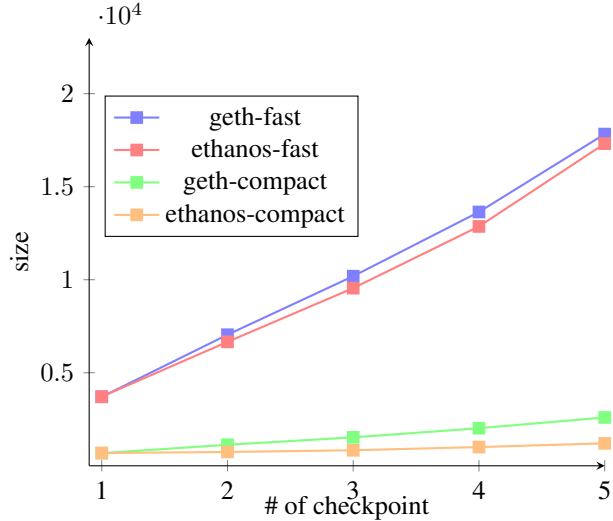

\subsubsection{Fast Sync and Compact Sync}
Fast sync downloads all the transactions, but replays only the transactions after the pivot block by downloading the pivot block state. Compact sync is same as fast sync except not downloading transactions before the pivot block. Table~\ref{table:sync_storage_fast_compact} shows storage size of geth and ethanos for each bootstrapping mode. Ethanos reduces \emph{Trie nodes} size of geth from 2,220MB to 804.79MB; however, it increases the transaction size (\emph{Bodies}, \emph{Receipts}) with block headers (\emph{Headers}). Therefore, ethanos did not reduce the size of fast sync much, but it significantly reduced compact sync from 2,592.54MB of geth to 1,206.97MB, which is less than 50\%. Our result shows that the trie node size of each checkpoint is about 800MB, so that if it is closer to the next checkpoint, compact sync size of ethanos could be increased up to 1,206.97 + 800 = 2,000.97MB, because it downloads both state tries in the last checkpoint and the pivot block, which is still smaller than geth compact sync.
Figure~\ref{fig:sync_size_of_each_checkpoint} is size of fast sync and compact sync between geth and ethanos every checkpoint showing the trends. We can see that the results are similar to the Figure~\ref{table:sync_storage_fast_compact}, and ethanos compact sync is getting more efficient over time, whereas ethanos full sync is similar to geth full sync. As ethanos compact sync confines the size of the state trie and transactions, the slight gradient is mostly caused by the increasing block headers.

\begin{figure*}
    \centering
    \begin{adjustbox}{width=\textwidth}
    \begin{tikzpicture}
        \begin{axis}[
            ybar, axis on top,
            height=7cm, width=15.5cm,
            bar width=0.3cm,
            ymajorgrids, tick align=inside,
            major grid style={dashed,draw=gray!50},
            enlarge y limits={value=.1,upper},
            ymin=0, ymax=8000,
            axis x line*=bottom,
            axis y line*=right,
            y axis line style={opacity=0},
            tickwidth=0pt,
            enlarge x limits=true,
            x tick label style={rotate=0, anchor=west, yshift=-0.2cm, font=\small},
            legend style={
                at={(0.5,-0.1)},
                anchor=north,
                legend columns=-1,
                font=\small,
                /tikz/every even column/.append style={column sep=0.5cm}
            },
            ylabel={Bootstrapping time (sec)},
            symbolic x coords={
               1,2,3,4,5},
           xtick=data,
        ]
        \addplot+[draw=none, fill=blue!30, error bars/.cd, y dir=both, y explicit] coordinates {
            (1,981.53) +- (0.0, 205.7875475)
            (2,2083.53) +- (0.0, 447.4637625)
            (3,2404.21) +- (0.0, 574.242649)
            (4,3779.05) +- (0.0, 1045.100472)
            (5,7067) +- (0.0, 280.9277996)};
        \addplot+[draw=none,fill=red!30, error bars/.cd, y dir=both, y explicit] coordinates {
            (1,1028.73) +- (0.0, 289.309638)
            (2,1755.93) +- (0.0, 296.5237805)
            (3,2536.8) +- (0.0, 718.9411063)
            (4,3328.07) +- (0.0, 900.1448666)
            (5,3697.67) +- (0.0, 1099.394747)};
        \addplot+[draw=none, fill=green!30, error bars/.cd, y dir=both, y explicit] coordinates {
            (1,704.25) +- (0.0, 263.825321)
            (2,1636.67) +- (0.0, 379.9052513)
            (3,2158.82) +- (0.0, 253.968806)
            (4,2929.11) +- (0.0, 245.8418405)
            (5,4351.06) +- (0.0, 467.651509)};
        \addplot+[draw=none, fill=orange!30, error bars/.cd, y dir=both, y explicit] coordinates {
            (1,666.53) +- (0.0, 186.0583114)
            (2,1066.93) +- (0.0, 151.1949861)
            (3,1550.8) +- (0.0, 218.1991882)
            (4,2055.47) +- (0.0, 240.1698407)
            (5,2985.19) +- (0.0, 1145.202993)};
        \legend{geth-fast,ethanos-fast,geth-compact,ethanos-compact}
        \end{axis}
    \end{tikzpicture}
    \end{adjustbox}
    \caption{Bootstrapping time of \textit{fast sync} and \textit{compact sync} between \emph{geth} and \emph{ethanos} at each checkpoint}
    \label{fig:bootstrapping_time}
\end{figure*}
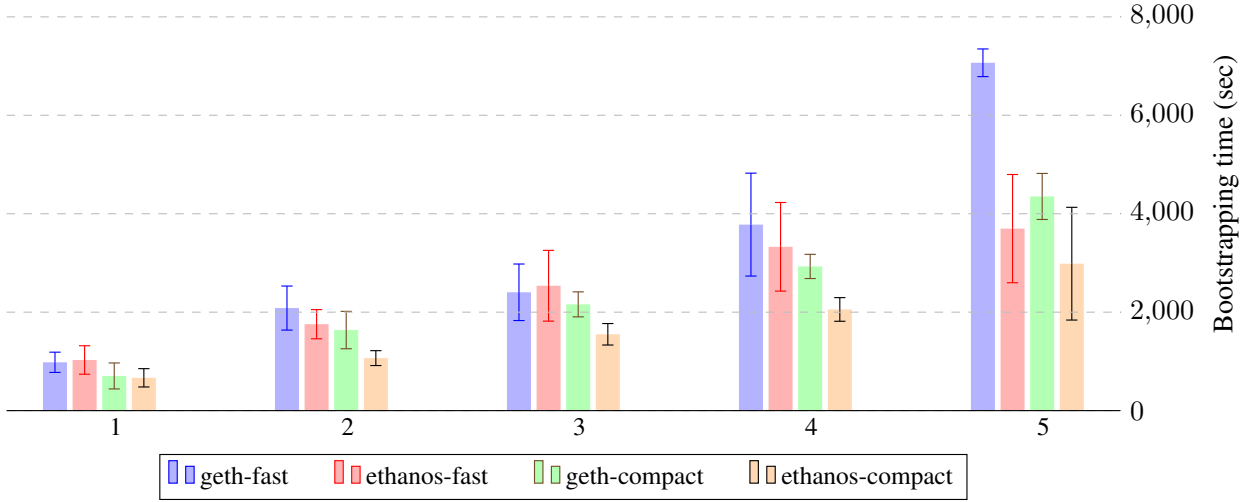

\subsection{Bootstrapping time}
Bootstrapping time is divided into \textit{download phase} and \textit{replay phase}, and download phase is more important between them, because it does not take much time to replay 64 blocks. Download phase could be divided again into state trie download phase and transaction download phase, which could be parallelized. Between them, state trie download is more important because transactions cannot be replayed until state trie download is completed. Therefore, state trie download phase is the primary bottleneck of bootstrapping. Figure~\ref{fig:bootstrapping_time} represents bootstrapping time of the four cases. As the checkpoints increase, all the cases has increased; however, the gradient of ethanos is smaller than geth in both fast sync and compact sync, because the growth of state trie of ethanos is confined, whereas that of geth dows not stop. Therefore, from the $5^{th}$ checkpoint, even the bootstrapping time for fast sync of ethanos begins to be smaller than that of ethereum compact sync.

\subsection{Overhead of restore transaction}
\begin{figure*}[ht]
\begin{adjustbox}{width=\linewidth}
\subfigure[Size of the normal and restore transactions.] {
\label{subfig:size_of_rstx}
\begin{tikzpicture}
    \begin{axis}[
        xlabel={checkpoint},
        ylabel={size (KB)},
        legend style={
            at={(0.5,-0.2)},
            anchor=north,
            legend columns=2,
            /tikz/every even column/.append style={column sep=0.5cm}
        },
        ymajorgrids=true,
        xmajorgrids=true,
        minor y tick num= 1,
        xtick={2,3,4,5},
        yminorgrids = true,
        minor grid style=loosely dotted,
        scatter/classes={
            normal={mark=*,blue!50,fill=blue!20},
            restore={mark=x,red!50}
        },
        only marks,
        scatter,
        mark size=3pt,
        scatter src=explicit symbolic]
        \addplot[
            color=blue!50,
            fill=blue!20,
            mark=*
        ] table[x=x,y=y]{tx_nor_size.dat};
        \addlegendentry{normal Tx}
        \addplot[
            color=red!60,
            mark=x
        ]
        table[x=x,y=y]{tx_res_size.dat};
        \addlegendentry{restore Tx}
    \end{axis}
\end{tikzpicture}
}
\subfigure[Execution time of the normal and restore transactions.] {
\label{subfig:time_of_rstx}
\begin{tikzpicture}
    \begin{axis}[
        xlabel={checkpoint},
        ylabel={time (ms)},
        legend style={
            at={(0.5,-0.2)},
            anchor=north,
            legend columns=2,
            /tikz/every even column/.append style={column sep=0.5cm}
        },
        ymajorgrids=true,
        xmajorgrids=true,
        minor y tick num= 2,
        xtick={2,3,4,5},
        yminorgrids = true, 
        minor grid style=loosely dotted,
        scatter/classes={
            normal={mark=*,blue!50,fill=blue!20},
            restore={mark=x,red!50}
        },
        only marks,
        scatter,
        mark size=3pt,
        scatter src=explicit symbolic,]
        \addplot[
            color=blue!50,
            fill=blue!20,
            mark=*
        ] table[x=x,y=y]{tx_nor_time.dat};
        \addlegendentry{normal Tx}
        \addplot[
            color=red!60,
            mark=x
        ]
        table[x=x,y=y]{tx_res_time.dat};
        \addlegendentry{restore Tx}
    \end{axis}
\end{tikzpicture}
}
\subfigure[The number of the normal and restore transactions] {
\label{subfig:number_of_rstx}
\begin{tikzpicture}
    \begin{axis}[
        title={\# of transactions},
        xlabel=\# of block,
        xtick={400000, 500000, 600000, 700000, 800000, 900000, 1000000},
        xmin=400000,
        xmax=1000000,
        ylabel = \# of transactions,
        legend style={
            at={(0.5,-0.2)},
            anchor=north,
            legend columns=2,
            /tikz/every even column/.append style={column sep=0.5cm}
        },
        x label style={at={(axis description cs:0.5,-0.1)},anchor=north},
        y label style={at={(axis description cs:-0.1,.5)},rotate=0,anchor=south},
        stack plots=y,
        area style,
    ]
    \addplot table [x=block,y=normal] {tx.dat} \closedcycle;
    \addlegendentry{normal Tx}
    \addplot table [x=block,y=restore] {tx.dat} \closedcycle;
    \addlegendentry{restore Tx}
    \end{axis}
\end{tikzpicture}
}
\end{adjustbox}
\caption{Size of the normal and Restore transactions.}
\label{fig:compare_transactions}
\end{figure*}
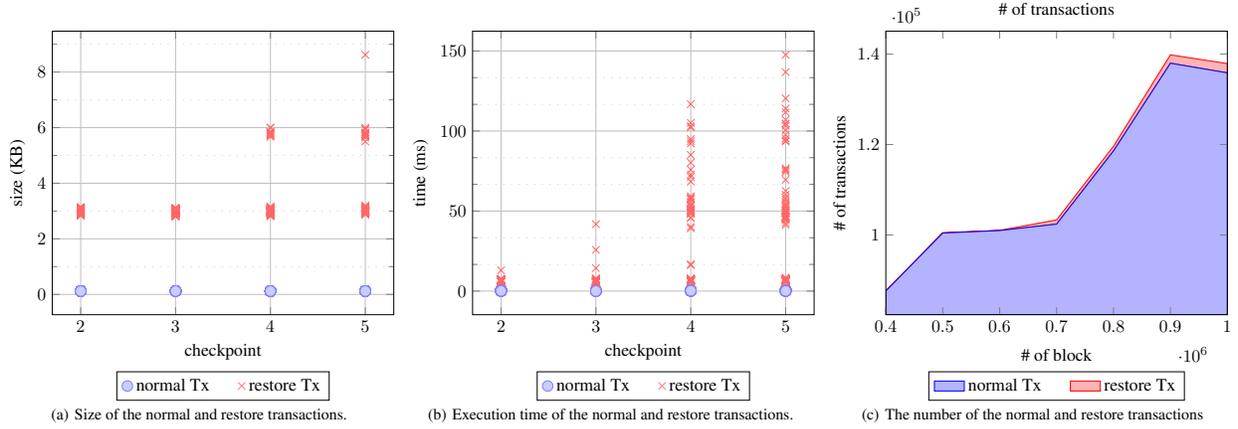

To see the overhead of restore transaction, we selected about 100 normal transactions and restore transactions each for every checkpoint. Figure~\ref{subfig:size_of_rstx} shows sizes of normal and restore transactions. The size of normal transactions is 0.12KB, but that of restore transactions varies over 3KB. Since every restore transaction has to include at least one merkle proof of which size is about 3KB that represents the state of the last active checkpoint. As the proof increases, the restore transaction size increases about 3KB for each proof.
Figure~\ref{subfig:time_of_rstx} depicts execution time of normal and restore transactions. The execution time for normal transaction is under 1 millisecond on average, whereas, the size of restore transactions is much larger then normal transactions due to the search time for bloom filters for each checkpoint block, and validation time for merkle proofs and void proofs. It takes about 150 ms for a restore transaction including 3 merkle or void proofs. Therefore, a miner is not willing to execute many restore transactions, which may raise the restore transaction fee.
Figure~\ref{subfig:number_of_rstx} is the number of normal and restore transactions from $7.04M$ to $8M$ blocks. The start block number is $7.04M$, because the accounts that appeared in the $1^{st}$ epoch are first dormant after $2^{nd}$ checkpoint $7,345,600$. The result represents that the normal transactions are increasing with the restore transactions. However, the number of restore transactions are much smaller than the normal ones less than 1.5\% at the every epochs. 

\section{Related Work} \label{sec:related_work}
There have been several \textit{ethereum improvement proposals (EIPs)} trying to clear state trie by removing accounts with negligible balance, which are \textbf{EIP-158}~\cite{eip158}, \textbf{EIP-161}~\cite{eip161} and \textbf{EIP-168}~\cite{eip168}. EIP-158~\cite{eip158} defined \textit{an empty account} of which \emph{nonce} and \emph{balance} are zero, also \emph{code} and \emph{storage} are empty. EIP-158 allows miners to delete empty accounts to save the storage, and EIP-161~\cite{eip161} covers several edge cases of EIP-158, which are included to \textit{Spurious Dragon} hard fork update~\cite{spurious_dragon}. EIP-168~\cite{eip168} suggested more aggressive optimization that removes \textit{dust accounts} of which balance is less than transaction fee, but it is still under discussion because of security issues.

\textbf{Vault}~\cite{leung2019vault} is a new cryptocurrency design based on \textit{Algorand}~\cite{gilad2017algorand} that minimizes storage and bootstrapping costs. Vault analyzed ethereum and found that 38\% of all accounts have no balances and no code or storages data. However, ethereum cannot remove those accounts because of replay attack. Vault solve the problem by introducing expiration date to transactions so that it can safely remove accounts of which balance is zero and the last transaction is expired. The experimental result shows that the state size is about 3.1GB with 500 million transactions when the state size of ethereum is about 5GB.

However, those EIPs and Vault targeted accounts with negligible balance, which do not required to be restored after the deletion because they are not worth to be retained. Whereas ethanos distinguishes accounts relatively less state-changing, but also worthless accounts, and temporarily deletes them like in hibernation. Therefore we could decrease more state size by removing more accounts, consequently synchronizing much faster.


\textbf{Utreexo}~\cite{dryjautreexo} is an optimization technique of Bitcoin by summarizing the UTXO set which is similar to states trie in ethereum. They found that 40\% of UTXOs last for 20 blocks or less and introduced a hash based accumulator to locally represent the UTXO set to compress the size of the set by logarithmic scale. With this method, users can spend a UTXO with a merkle proof, and miners can validate it with the accumulator. They have yet to measure the whole bootstrapping cost, but simulated download size of Bitcoin’s blockchain up to early 2019 with 500MB cache showing that Utreexo greatly reduces the proof size with sufficient cache memory.

\textbf{FlyClient}~\cite{bunz2019flyclient} proposed an ethereum light client that overcomes limitations of NIPoPoW~\cite{kiayias2017non}. FlyClient supports blockchains of variable difficulty, which NIPoPoW does not support, and suggested an optimal probablistic block sampling protocol and \textit{merkle mountain range (MMR)} commitments, which significantly reduces the proof size compared to SPV proofs. FlyClient reduces the size of light client, however, light client is not a full node that can verify a transaction without help of other nodes. Compact sync of ethanos has a same problem with light clients because block header is linearly increasing, so that if we apply the idea of FlyClient, we will be able to reduce the size of compact sync further.

\section{Conclusion} \label{sec:summary}
Ethanos has reduced bootstrapping cost by sweeping dormant accounts and old transactions backed up by restoration mechanism and incentive structure. We experimented our work with real ethereum data from $7M+1$ to $8M$ blocks and found that ethanos significantly reduces storage size and bootstrapping time, which enables ordinary clients using PCs or smartphones to bootstrap efficiently to send or verify a transaction themselves.

In some cases, however, a restore transaction would require large payload data for merkle proofs and void proofs, which cost a lot of transaction fee. This problem could be solved because the users will use ethanos in an efficient way if they understand the mechanism of ethanos. For example, paying a storage fee periodically or using a bank service that provides custody or loan business that demands low fees or pays interest.

Ethanos is an optimization technique of ethereum, but it is not limited to ethereum and could be applied to any account-based blockchain that uses MPT. In addition, we did not treat smart contracts for simplicity in this paper, and left them as a future work.

\section*{Acknowledgement}
We thank Heewon Chung, Jun-Woo Choi, Duekeun Kim, Ji-Young Um, Seo-Young Ko, Jin-Young Yoo, and Georae Han, members of Decipher, blockchain research group at Seoul National University, who initially helped develop the idea.

\bibliographystyle{unsrt}  


\end{document}